%% file: ms.tex
\pdfoutput=1 
\documentclass{svjour3}

\usepackage{pdflscape}
\usepackage{pdfpages}

\usepackage{graphicx}
\usepackage{amssymb}
\usepackage{amsmath}
\usepackage{algorithm2e}
\usepackage[caption=false]{subfig}

\usepackage{tikz}
\usetikzlibrary{decorations.pathreplacing}
\usetikzlibrary{fit}
\usetikzlibrary{calc}
\usetikzlibrary{positioning}
\usetikzlibrary{arrows.meta}
\usetikzlibrary{shapes}
\usepackage{float}

\newcommand{\citationneeded}[1][]{\textsuperscript{[citation needed]}}

\newcommand{\gscalar}[1]{\MakeLowercase{{#1}}}
\newcommand{\gvector}[1]{\mathbf{\MakeLowercase{{#1}}}}
\newcommand{\gmatrix}[1]{\mathbf{\MakeUppercase{{#1}}}}
\newcommand{\gtensor}[1]{\mathsf{\MakeUppercase{{#1}}}}
\newcommand{\gset}[1]{\mathbb{\MakeUppercase{{#1}}}}

\newcommand{\ninputs}{{n_i}}
\newcommand{\nhiddens}{{n_h}}
\newcommand{\noutputs}{{n_o}}
\newcommand{\nneurons}{n}
\newcommand{\nlayers}{{n_l}}
\newcommand{\nticks}{{n_t}}
\newcommand{\iho}{iho\;}

\newcommand{\currentstate}{{ \gvector{S_n} }}
\newcommand{\previousstate}{{ \gvector{S_{n-1}} }}
\newcommand{\presynaptic}{{ \gvector{T_n} }}

\newcommand{\currentstatecomp}{{ s_{n,o} }}
\newcommand{\presynapticcomp}{{ t_{n,o} }}
\newcommand{\previousstatecomp}{{ s_{n-1,k} }}

\begin{document}

\title{Formal derivation of Mesh Neural Networks with their Forward-Only gradient Propagation}

\titlerunning{Mesh Neural Networks with their Forward-Only gradient Propagation}

\author{Federico A. Galatolo \and Mario G.C.A. Cimino \and Gigliola Vaglini}
\institute{Department of Information Engineering, University of Pisa, 56122 Pisa, Italy \email{federico.galatolo@ing.unipi.it, mario.cimino@unipi.it, gigliola.vaglini@unipi.it}}

\begin{landscape}
\includepdf[pages=-, angle=90]{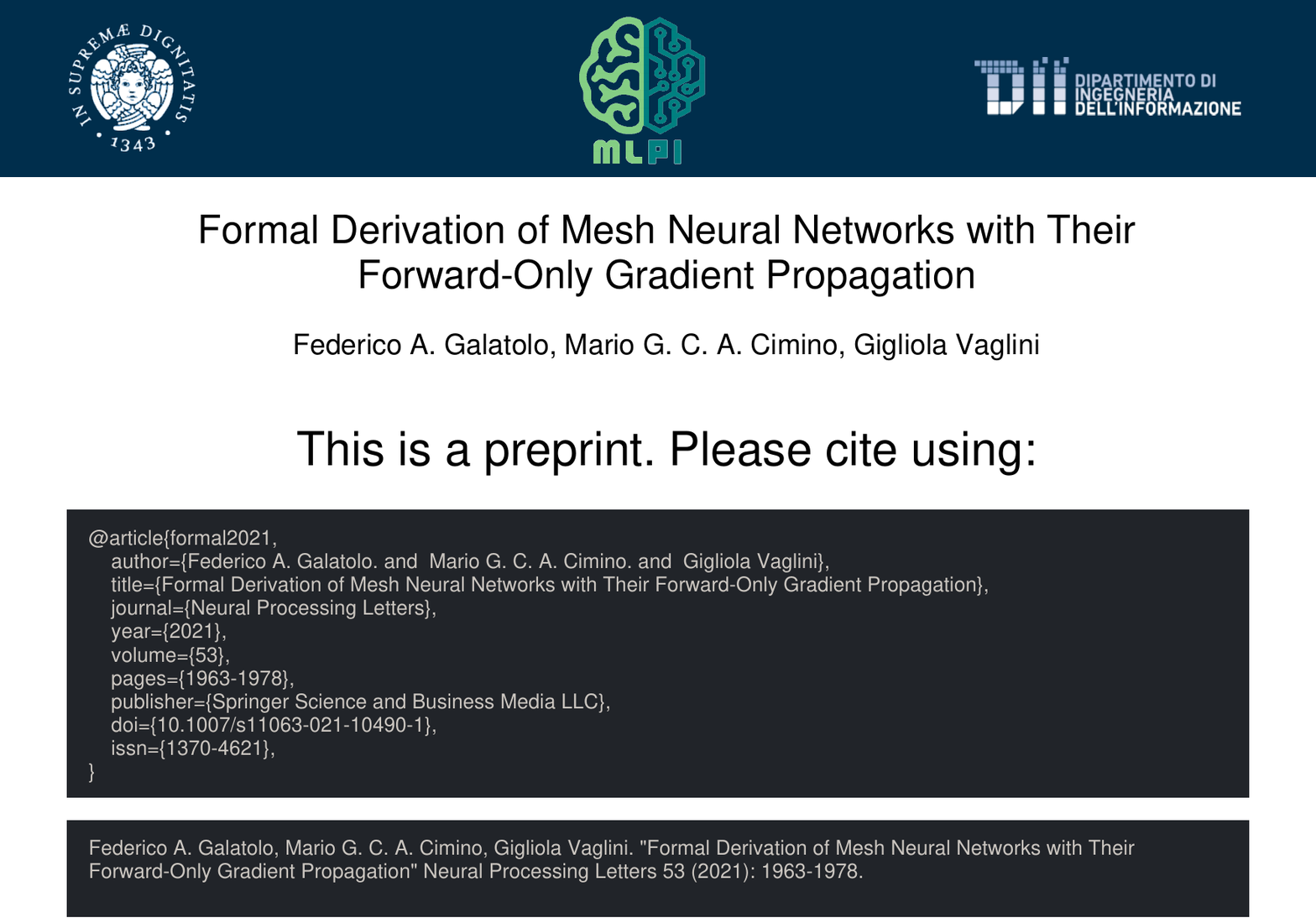}
\end{landscape}

\maketitle

\begin{abstract}
This paper proposes the Mesh Neural Network (MNN), a novel architecture which allows neurons to be connected in any topology, to efficiently route information. In MNNs, information is propagated between neurons throughout a state transition function. State and error gradients are then directly computed from state updates without backward computation. The MNN architecture and the error propagation schema is formalized and derived in tensor algebra. The proposed computational model can fully supply a gradient descent process, and is potentially suitable for very large scale sparse NNs, due to its expressivity and training efficiency, with respect to NNs based on back-propagation and computational graphs.
\keywords{Artificial Neural Networks \and Gradients Computation \and Supervised Learning \and Deep Learning}
\end{abstract}

\section{Introduction and background}

A huge amount of research has been made during the last years on a variety of applications of Artificial Neural Networks (ANNs). As a consequence, many ANNs architectures have been developed, generating surrogate models from different types of big data, such as image, audio, video, text, time series, and so on. With ANNs, the underlying relationships among data can be approximated with little knowledge of the system to be modelled. In spite of this success, ANNs are computational models vaguely inspired to biological brains, and require relevant computation and management with respect to the biological counterpart.

Specifically, Deep Learning is achieving good levels of performance, via architectures composed of several layers. The Deep Learning research is mostly based on gradient-based optimization methods and on the well-known \textit{backpropagation} (BP) algorithm. In essence, BP includes a forward and backward layer-wise computation of the loss function with respect to the neurons weights. Actually, BP is not biologically plausible. Moreover, convergence problems, such as vanishing and exploding gradients, occur when using many layers. Finally, BP can be very unstable when dealing with recurrent networks and can be ineffective to exploit long-lasting relationships \cite{Galatolo_2019}.
In the last decade, an increasing number of alternative strategies have proposed to simplify the ANN training. A first strategy consists in removing the backward computation by deriving a forward only computation. A reference work for this approach is \cite{Wilamowski2010}. Specifically, the proposed method improves the efficiency of Jacobian matrix computation, for fully or partially connected ANNs. An interesting advantage of this approach is that
it can train arbitrarily connected ANNs, and not just Multi-Layer Perceptron (MLP)-based architectures. Indeed, ANNs with connections across layers are much more powerful than MLPs. A more recent research in which the Jacobian matrix is calculated only in the forward computation was made by Guo \textit{et al.} \cite{Guo2017}. In general, to remove the backward computation is not costless: an additional calculation in the forward computation must be considered. However, the forward-only computation is more parallelizable than traditional forward and backward computations, as the dataset is large and the number of hidden neurons increases. A different strategy is presented in  \cite{ma2019hsic}, in which the training method is based on a different principle called information bottleneck, which does not require backpropagation. In general, a performance comparison with BP is difficult, since performance can heavily depend on the minibatch size. The minibatch size is usually a constant that is based on available GPU memory. On the other side, a quantity of interest is the learning convergence, which is unknown for either BP or other methods. Since the backward computation is removed for such approaches, they are more suitable for parallel computation. Another type of strategy is proposed by Jaderberg \cite{jaderberg2017decoupled}: a model for predicting gradient, called synthetic gradient, is calculated in place of true backpropagation error gradients. With such synthetic gradients, layers can be independently updated, removing forward and update locking.

According to this research trend, this paper formally introduces recent advances leading to a novel, arbitrarily connected, ANNs architecture, in which error gradients are computed throughout a state transition function without backward computation. The paper is organized as follows. In Section 2, the fundamentals of the problem are defined. A formal derivation of the proposed architecture is presented in Section 3. Section 4 covers the implementation and experimental aspects. Section 5 is devoted to conclusions and future work.

\section{Problem statement}

An Artificial Neurons Layer (ANL) with $\ninputs$ inputs and $\noutputs$ outputs can be described by its \textit{layer weights matrix} $\gmatrix{W} \in \gset{R}^{\ninputs \times \noutputs}$ and activation function $\hat{\varphi}(\gvector{x}): \gset{R}^\noutputs \rightarrow \gset{R}^\noutputs$. Let us consider activation functions for which it holds that $\hat{\varphi}(\gvector{x})_i = \varphi(x_i)$ (where $\varphi(x) : \gset{R} \rightarrow \gset{R}$). Each column $\gmatrix{W}_{*,i}$ of $\gmatrix{W}$ represents the weights vector from the inputs to the $i$-th perceptron, in which biases are represented as weights of fictitious inputs that always produce the constant value 1. Given the input vector $\gvector{x} \in \gset{R}^\ninputs$, the output vector $\gvector{y} \in \gset{R}^\noutputs$ of the ANL is $\gvector{y} = \varphi(\gvector{x}\gmatrix{W})$. In multilayer neural networks, or MLPs, ANLs are stacked, i.e., the ANL$_i$ is fed by the output of the ANL$_{i-1}$: each set of weights connecting the $i$-th layer is represented by a different matrix $\gmatrix{W}_i$, and the input/output layers are considered as special topological elements with respect to the hidden layers.

In the popular BP training algorithm, the gradients of the weights are iteratively computed exploiting a propagation rule between layers \cite{Keller2016,bp_werbos}. Let us consider a generic error function $E(\gvector{y}, \gvector{\overline{y}}): \gset{R}^{\nneurons\times \gscalar{2}} \rightarrow \gset{R}$ that computes the error between a network output $\gvector{y}$ and a desired one $\gvector{\overline{y}}$, and a generic error function with respect to the $o$-th output $y_o$ $E_o(y_o, \overline{y_o}): \gset{R}^2 \rightarrow \gset{R}$. Let us assume that $E(\gvector{y},\gvector{\overline{y}})$ is a composition of $E_o(y_o, \overline{y_o})$ for every output unit.
Considering an MLP with $\nlayers$ layers, the objective of the BP algorithm is to compute the gradients of every output error $\frac{\partial E(y_o,\overline{y_o})}{\partial p_i}$ with respect to every parameter $p_i$.
Such gradients can be used by a \textit{Stochastic Gradient Descent} (SGD) algorithm to train the MLP \cite{Theodoris2015}. Let $net_{i,o}$ be the $o$-th output of the $i$-th hidden layer. Applying the  chain rule for differentiating composite functions to $\frac{\partial E(y_o,\overline{y_o})}{\partial p_i}$, the corresponding error gradient is:
\begin{flalign}
\frac{\partial E(y_o,\overline{y_o})}{\partial p_i} = \frac{\partial E(y_o,\overline{y_o})}{\partial y_o}\frac{\partial y_o}{\partial p_i} = \frac{\partial E(net_{L-1,o},\overline{y_o})}{\partial net_{L-1,o}}\frac{\partial net_{L-1,o}}{\partial p_i}.
\end{flalign}
The derivative $\frac{\partial E(net_{\nlayers-1,o},\overline{y_o})}{\partial net_{\nlayers-1,o}}$ depends on the error function and is known. In the derivative $\frac{\partial net_{\nlayers-1,o}}{\partial p_i}$, each parameter of a layer influences the output values of all the subsequent layers. Hence, in order to compute $\frac{\partial net_{\nlayers-1,o}}{\partial p_i}$, the chain rule is applied up to the term $\frac{\partial net_{i,o}}{\partial p_i}$. For this purpose, the BP algorithm iteratively applies the chain rule on each layer in reverse order for efficiently computing the partial derivatives with respect to all parameters. More formally, given the output of the $l$-th layer, $\gvector{net_l} = \varphi(\gvector{net_{l-1}}\gmatrix{W}_l)$, let us say its $o$-th element $t_{l,o} = (\gvector{net_{l-1}}\gmatrix{W}_l)_o$. The chain rule is applied to $\varphi(t_{l,o})$, and in order to compute the term $\frac{\partial \varphi(t_{l,o})}{\partial t_{l,o}}$, $\gvector{t_l}$ needs to be saved for each layer.

To train ANNs without a layered topology, the approach commonly used is the automatic differentiation on \textit{computational graphs} (CGs) \cite{goodfellow}, in which computations are represented in a graph. In essence, for each operation (e.g., matrix multiplication, element-wise sum, etc.) the inputs $x_0, x_1, \cdots, x_{n-1}$ and the output $\gvector{y}$ are represented as incoming and outgoing edges of a graph, respectively. For each edge $\frac{\partial y}{\partial x_i}$ is computed. For a given ANN, the operations to compute its output $y_o$ and the error $E(y_o,\overline{y_o})$ are then represented as a CG. Let us consider, a ``factoring path'', i.e., a path between two nodes in which the derivatives $\frac{\partial y}{\partial x_i}$ encountered on the traversed edges are all multiplied together. Then, the partial derivative of the error function with respect to a parameter, i.e., $\frac{\partial E(y_o,\overline{y_o})}{\partial p_i}$, is the sum of all the reverse factoring paths from $E(y_o,\overline{y_o})$ to $p_i$, i.e., the paths belonging to the set $\mathcal{P}_i$:
\begin{flalign} 
\frac{\partial E(y_o,\overline{y_o})}{\partial p_i} = \sum\limits_{p \in \mathcal{P}_i}\prod_{(x,y) \in p }\frac{\partial y}{\partial x}.
\end{flalign}

A CG representation is a general formalism to represent all network topologies, such as feedforward, recurrent, convolutional, residual, and so on. To train arbitrarily connected ANNs topologies is very important, because ANNs with connections across layers are much more powerful than classical MLP architectures. However, a CG increases the space complexity with respect to a corresponding MLP-based representation (where an MLP representation is possible). Indeed, the underlying data structure needs to store both the graph topology and the partial derivatives $\frac{\partial y}{\partial x_i}$ of each edge. Moreover, it results in a higher time complexity, because all the reverse factoring paths have to be found.

In the next section, a novel ANNs representation is introduced, which is capable of training arbitrarily connected neural networks and, as a consequence, ANNs with reduced number of neurons and
good generalization capabilities. The interesting properties of the training algorithm is the lack of a backpropagated computation, and an iteration without need of memory relationships than the one with the previous step. Hence, the  proposed method is much simpler than traditional forward and backward procedure. Indeed, the training iteration can be described by three matrix operations. Due to the possibility of training unstructured ANNs, the proposed architectural model is called Mesh Neural Network (MNN).

\section{Formal derivation of a Mesh Neural Network}

\subsection{Structure, activation and state of an MNN}
The proposed MNN is based on a matrix representation that is not a transfer matrix, but it is an \textit{adjacency matrix} (AM), i.e., a square matrix representing the ANN as a finite graph. The elements of the AM indicate whether pairs of vertices are adjacent or not in the graph, by means of a non-zero or zero weight, respectively.

More formally, an AM $\gmatrix{A}$ is a matrix in which each element $A_{i,j}$ represents the weight from the node $i$ to the node $j$. For example MLPs are a subset of the representable topologies with AMs: since in MLPs only connections between layers are possible, their AMs are block matrices. Figure \ref{fig:ff_adj} shows an MLP topology with the corresponding AM. Here, each $\gmatrix{W}_i$ is the weights matrix of the $i$-th layer and occupies a corresponding block in the AM.

\begin{figure}[!ht]
\subfloat[ANN Topology\label{fig:ff}]{%
\begin{minipage}[c][1\width]{0.5\textwidth}%
\input{figures/ff.tex}
\end{minipage}}
\subfloat[Adjacency Matrix\label{fig:adj_ff}]{
\centering{}%
\begin{minipage}[c][1\width]{0.5\textwidth}%
\begin{center}
\input{figures/adj_ff.tex}
\par\end{center}%
\end{minipage}}
\caption{An MLP and its adjacency matrix}
\label{fig:ff_adj}
\end{figure}

An example of unstructured topology and its corresponding AM is shown in Figure \ref{fig:am_adj}.

\begin{figure}[!ht]
\subfloat[ANN Topology]{%
\begin{minipage}[c][1\width]{0.5\textwidth}%
\input{figures/am.tex}
\end{minipage}}
\subfloat[Adjacency Matrix]{
\centering{}%
\begin{minipage}[c][1\width]{0.5\textwidth}%
\begin{center}
\input{figures/adj_am.tex}
\par\end{center}%
\end{minipage}}
\caption{An unstructured ANN and its adjacency matrix}
\label{fig:am_adj}
\end{figure}

A generic MNN topology with $\nneurons$ neurons is represented by a matrix $\gmatrix{A} \in \gset{R}^{\nneurons\times \nneurons}$. It is worth noting that this representation does not include the topological distinction between input, hidden and output neurons. Let $\ninputs$,$\noutputs$, and $\noutputs$ be the number of input, hidden and output neurons. Since all neurons are identified by a position in the matrix, a good convention (hereinafter called ``\iho positioning convention'') to distinguish the three sets without loss of generality is to assign them a positioning: to consider the first $\ninputs$ elements as input neurons, the subsequent $\nhiddens$ elements as hidden neurons, and the last $\noutputs$ elements as output neurons.

Let the \textit{state} be $\gvector{s_i} \in \gset{R}^\nneurons$ the output value of each neuron in the MNN at the $i$-th instant of time. The output of an MNN is provided along a temporal sequence, whose length depends on the distances between input and output neurons. This allows an MNN to exhibit temporal dynamic behavior. Let us recall that: (i) $A_{i,j}$ represents the weight from neuron $i$ to neuron $j$; (ii) the $h$-th neuron output is computed as $\varphi(\sum_{k=0}^N w_{k,h} x_k)$; (iii) biases are represented as weights of fictitious inputs that always produce the constant value 1. Hence, given an initial state $\gvector{s_0}$, which is set to the input value for input neurons and to zero for the other neurons, the next state is calculated as:
\begin{flalign}
\label{eq:updateRule}
\gvector{\currentstate} = \hat{\varphi}(\gvector{\previousstate }\gmatrix{A})
\end{flalign}

At each time tick, the state transition of each neuron can influence the outputs values of all adjacent neurons. For subsequent ticks, the initial piece of information contained in $\gvector{s_0}$ can traverse subsequent neurons and can influence their states, up to the output neurons.
\subsection{Derivation of state and error gradients}

In this section, the error derivative $\frac{\partial E(y,\overline{y})}{\partial p_i}$ for every parameter $p_i$ of an MNN are formally determined. It can be observed from Equation (\ref{eq:updateRule}) that the unique parameter is $\gmatrix{A}$. Let us assume an MNN with $\nneurons$ neurons, of which $\ninputs$ input neurons and $\noutputs$ output neurons positioned in the matrix according to the \iho ordering convention. Let be the MNN processed for $t$ states. 
The $o$-th output value is then $y_o = s_{t-1,o} = \hat{\varphi}(\gvector{s}_{t-2}\gmatrix{A})_o$ where $o \in \{\nneurons-\noutputs,\cdots,\nneurons-1\}$. Recalling the chain rule:
\begin{flalign}
\frac{\partial E(y_o,\overline{y_o})}{\partial p_i} = \frac{\partial E(y_o,\overline{y_o})}{\partial y_o} \frac{\partial s_{t-1,o}}{\partial p_i}. \label{eq:errGrad}
\end{flalign}

Let us consider a generic state $\currentstate = \hat{\varphi}(\presynaptic  )$ where $\presynaptic   = \previousstate \gmatrix{A}$. According to the chain rule, the derivative for a generic output $o$ is:
\begin{flalign}
\frac{\partial \currentstatecomp  }{\partial A_{i,j}} = \frac{\partial\varphi(\presynapticcomp   )}{\partial \presynapticcomp   } \frac{\partial \presynapticcomp   }{\partial A_{i,j}} = \frac{\partial\varphi(\presynapticcomp   )}{\partial \presynapticcomp   } \frac{\partial (\previousstate  \gmatrix{A})_o}{\partial A_{i,j}} \label{eq:diffSn}
\end{flalign}
where $(\previousstate \gmatrix{A})_o$ is:
\begin{flalign}
(\previousstate \gmatrix{A})_o = \sum\limits_{k=0}^N \previousstatecomp A_{k,o} \label{eq:summation}
\end{flalign}

Let us distinguish two cases in Equation (\ref{eq:summation}): (i) if $o=j$, one of the $A_{k,o}$ is $A_{i,j}$; (ii) if $o\neq j$, all the $A_{k,o}$ are constant with respect to $A_{i,j}$. Let us consider the case $o = j$. For linearity of differentiation: 

\begin{flalign}
\frac{\partial (\previousstate \gmatrix{A})_j}{\partial A_{i,j}} = \frac{\partial(\sum\limits_{k=0}^N \previousstatecomp A_{k,j})}{\partial A_{i,j}} = \sum\limits_{k=0}^N\frac{\partial(\previousstatecomp A_{k,j})}{\partial A_{i,j}} \label{eq:am_summation}
\end{flalign}

In the partial derivatives $\frac{\partial(\previousstatecomp A_{k,j})}{\partial A_{i,j}}$, all the $\previousstatecomp $ elements depend on $A_{i,j}$. Moreover, in the case $k \neq i$, the matrix elements $A_{k,j}$ are constants with respect to $A_{i,j}$. Let us distinguish in Equation (\ref{eq:am_summation}) the term with $k = i$:

\begin{flalign}
&\sum\limits_{k=0}^N\frac{\partial(\previousstatecomp A_{k,j})}{\partial A_{i,j}}=\sum\limits_{k=0,\;k\neq i}^N\frac{\partial(\previousstatecomp A_{k,j})}{\partial A_{i,j}} + \frac{\partial(s_{n-1,i}A_{i,j})}{\partial A_{i,j}}\label{eq:summation2}
\end{flalign}

Since $A_{k,j}$ is a constant, the first term of Equation (\ref{eq:summation2}) is:

\begin{flalign}\label{eq:am_sum_diff}
&\sum\limits_{k=0,\;k\neq j}^N\frac{\partial(\previousstatecomp A_{k,j})}{\partial A_{i,j}} = \sum\limits_{k=0,\;k\neq j}^N \frac{\partial \previousstatecomp }{\partial A_{i,j}}A_{k,j}
\end{flalign}

By applying the product rule to the second term of Equation (\ref{eq:summation2}):

\begin{flalign}
&\frac{\partial(s_{n-1,i}A_{i,j})}{\partial A_{i,j}} = \frac{\partial s_{n-1,i}}{\partial A_{i,j}}A_{i,j} + \frac{\partial A_{i,j}}{\partial A_{i,j}}s_{n-1,i} = \frac{\partial s_{n-1,i}}{\partial A_{i,j}}A_{i,j} + s_{n-1,i}
\end{flalign}

The term $\frac{\partial s_{n-1,i}}{\partial A_{i,j}}A_{i,j}$ can be integrated in the summation of Formula (\ref{eq:am_sum_diff}):

\begin{flalign}
\sum\limits_{k=0}^N\frac{\partial(\previousstatecomp A_{k,j})}{\partial A_{i,j}} = \sum\limits_{k=0}^N \frac{\partial \previousstatecomp }{\partial A_{i,j}}A_{k,j} + s_{n-1,i}
\end{flalign}

Similarly, considering the case $o \neq j$ in Equation (\ref{eq:summation}), the $A_{k,o}$ elements are constant with respect to $A_{i,j}$, leading to:

\begin{flalign}
\frac{\partial (\previousstate \gmatrix{A})_o}{\partial A_{i,j}} = \frac{\partial(\sum\limits_{k=0}^N \previousstatecomp A_{k,o})}{\partial A_{i,j}} = \sum\limits_{k=0}^N\frac{\partial \previousstatecomp }{\partial A_{i,j}}A_{k,o}
\end{flalign}

Hence, Equation (\ref{eq:diffSn}) can be formulated as follows:

\begin{flalign}\label{eq:am_der}
\frac{\partial \currentstatecomp  }{\partial A_{i,j}} = \begin{cases}
\frac{\partial\varphi(\presynapticcomp   )}{\partial \presynapticcomp   }(\sum\limits_{k=0}^N \frac{\partial \previousstatecomp }{\partial A_{i,j}}A_{k,j} + s_{n-1,i}) & \text{if } o = j\\
\frac{\partial\varphi(\presynapticcomp   )}{\partial \presynapticcomp   }(\sum\limits_{k=0}^N \frac{\partial \previousstatecomp }{\partial A_{i,j}}A_{k,o}) & \text{if } o \neq j
\end{cases}
\end{flalign}

As a result, Equation (\ref{eq:am_der}) determines a very efficient algorithm for computing the partial derivative of the MNN state, which is, in turn, essential for applying an SGD-based training. In  three terms: (i) the partial derivatives of the activation function $\frac{\partial\varphi(\presynapticcomp   )}{\partial \presynapticcomp   }$, (ii) the previous states $\previousstatecomp $, and (iii) the partial derivatives previous state $\frac{\partial \previousstatecomp }{\partial A_{i,j}}$. Consequently, it is possible to compute both the next states $s_{n,o}$ and the next state partial derivatives $\frac{\partial s_{n,o}}{\partial A_{i,j}}$, concurrently and in the same iteration step. Moreover, an iteration does not need to store any intermediate values except for those of the current state, which can then be overwritten in the next iteration. Since the error gradient can be directly calculated from state gradient, Equation (\ref{eq:errGrad}) results in a simplified iterative method without any memory dependency than the one with the previous step.

Operations in Equation (\ref{eq:am_der}) can be performed with scalars, vectors, and matrices, and then can be reformulated so as to be efficiently performed with tensors. In the next section, Equation (\ref{eq:am_der}) and the error gradient propagation schema are formalized and derived by tensor algebra.

\subsubsection{Tensor Algebra formulation of the error gradient}

Let us denote by $\nabla_\gmatrix{A}\gvector{s_n} \in R^{N\times N\times N}$ the tensor of the partial derivatives $\frac{\partial \currentstatecomp  }{\partial A_{i,j}}$
\begin{flalign}
(\nabla_\gmatrix{A}\gvector{s_n})_{i,j,o} = \frac{\partial \currentstatecomp  }{\partial A_{i,j}}
\end{flalign}
and by $\nabla_\gvector{x} \varphi(\gvector{x})$ the tensor of partial derivatives $\frac{\partial \varphi(x_i)}{\partial x_i}$
\begin{flalign}
(\nabla_\gvector{x} \varphi(\gvector{x}))_i = \frac{\partial \varphi(x_i)}{\partial x_i}
\end{flalign}
and by $\gtensor{\tilde{S}}_n \in R^{N\times N\times N}$ a tensor such that:
\begin{flalign}
\gtensor{\tilde{S}}_{i,j,o} = \begin{cases}
s_{n,i} & \text{if } o = j\\
0       & \text{otherwise}
\end{cases}
\end{flalign}

Hence, it is possible to formulate Equation \ref{eq:am_der} as:

\begin{flalign}
\nabla_\gmatrix{A}\gvector{\currentstate} = \nabla_\gvector{\presynaptic} \varphi(\gvector{\presynaptic}) \odot (\nabla_\gmatrix{A}\gvector{\previousstate} \gmatrix{A} + \gtensor{\tilde{S}}_n)
\end{flalign}

where the symbol $\odot$ denotes the Hadamard product.

As a result, the error gradient Forward-Only Propagation (FOP) algorithm of an MNN can be formulated in terms of the following steps, i.e., initialization, state derivatives forward propagation, and error derivative computation:

\setcounter{figure}{0}
\renewcommand{\figurename}{Algorithm}
\begin{figure}[ht]
  \centering
  \begin{minipage}{.5\linewidth}
  \input{figures/am_algo.tex}
  \end{minipage}
  \caption{FOP algorithm for the error gradient of an MNN}
  \label{fig:algo}
\end{figure}

where $\nticks$ is the number of timesteps needed to the input to traverse the network and provide a sufficiently accurate output. In Recurrent Neural Networks (RNNs) a careful consideration is required to determine the value of $\nticks$, because any recurrent connection results in a potentially undefined number of loops. However, a relevant advantage of MNN with respect to RNN based on back-propagation is that an MNN does not need to save the prior steps determined by a loop. In RNNs a bounded-history approximation strategy is used to simplify the computation and provide an adequate approximation to the true gradient: relevant information is saved in the fixed number of timesteps $\nticks$  and any information older than that is forgotten. According to this strategy, in Backpropagation Through Time \cite{bprnn}, a backward pass through the most recent $\nticks$ time steps is performed at each time the network is run through an additional time step. In contrast, in MNN the lack of an error backpropagation sensibly reduces the impact of $\nticks$: it should be large enough to capture the temporal structure of the problem to model. Thus, after $\nticks$ timesteps the computation is simply truncated to take the output value. It is worth noting that already in the training phase weights are adjusted according to the specified $\nticks$. Consequently, recurrent connections are adequately weakened when producing noise on the error, reducing the impact of the recurrent computation.In conclusion, a sufficiently large $\nticks$  results in an adequate approximation to the true gradient, and it is not a sensitive parameter of the network.

The next section is devoted to the Python implementation and the evaluation of the proposed MNN.

\section{Implementation and experimental studies}

The MNN model has been developed, tested and publicly released on the Github platform, to make possible the initial roll-out of the approach, and to foster its application on various research environments.
The implementation is based on \textit{numpy}\cite{numpy}, a widespread package for tensor algebra in Python. The interested reader is referred to \cite{mnnrepo} for further implementation details.

The correctness of the symbolic derivatives is a critical aspect of the proposed network. To ensure it, in addition to the symbolic differentiation (SD), another implementation has been generated, in which gradients are calculated via automatic differentiation (AD) \cite{pytorch}. AD transforms a target function into a large graph of symbolic differentiation at elementary operation level, which are highly parallelizable \cite{autograd}. This computational graph can efficiently manage orders of magnitude of gradients, providing highly accurate numerical values. Nevertheless, the AD-based system can be used for testing purposes only, since it is based on a back-propagated gradient error that has been criticized in the premise of this research work. To develop an efficient and coherent implementation of the proposed approach, the symbolic derivatives are then fundamental.
To empirically evaluate the functional equivalence of the SD-based and the AD-based networks, the absolute differences between their corresponding output values have been computed over 100 tests. The two networks have been equipped with 5 input, 10 hidden and 3 output nodes. In each test, the two comparative networks have been set with (the same) random weights, and fed with a batch of (the same) 10 random inputs. As a result, the 95\% confidence intervals of both the state and the gradient absolute differences are very low: $0.00024 \pm 0.000047$ and $0.000011 \pm 0.0000053$, respectively. The source code of the numerical test code has been publicly released \cite{mnnrepo}.

\subsection{Synthetic problems }

In order to investigate the capabilities of the MNN model the dataset generator of \textit{scikit-learn}\cite{scikit-learn} has been used to produce five types of two-dimensional dataset well-known in the literature (Figures \ref{fig:2_class_exp}, \ref{fig:3_class_exp}, \ref{fig:iris}): (a) \textit{Moons}: a two-classes dataset made by two interleaving circles; (b) \textit{Circles}: a two-class dataset made by concentric circles; (c) \textit{Spirals}, which is considered as a good evaluation
of training algorithms \cite{Wilamowski2010}; (d) \textit{Single Blobs}: a three-class dataset made by isotropic Gaussian blobs with standard deviation 1.0, 2.5, 0.5; (e) \textit{Double Blobs}: a three-class dataset made by two groups of isotropic Gaussian blobs with standard deviation 1.0. 

Each dataset is made by 1,000 objects, balanced classes, and contains 10\% of noise. Finally, a dataset from UCI Machine Learning Repository has been used, known as Iris \cite{iris}. Iris contains three classes of Iris plants. Each class
consists of 50 objects characterised by 4 numeric features which describe, respectively, sepal length, sepal width,
petal length and petal width. Class Iris Setosa is linearly separable from the other two. However, class Iris Versicolor and Iris Viginica are not separable from each
other.

\setcounter{figure}{2}
The MNN topology represented in Figure \ref{fig:exp_mnn} has been used. Specifically, two output units have been assigned for the two classes datasets, and three output units for the three classes datasets. On the other side, three inputs units have been used: two inputs for the $(x,y)$ features of the dataset, and one input for the bias input (constantly set to  1). 5 hidden units have been used. The Network has been evaluated for 3 time ticks. The ReLU activation function has been used for all units. Finally, the cross-entropy loss has been used as error function.
For the experiments using the \textit{Iris} dataset, it has been used an MNN with 5 input units (4 features and 1 bias), 10 hidden units and 3 output units (one for each class). 
\\
\renewcommand{\figurename}{Figure}
\begin{figure}[ht]
  \centering
  \scalebox{.8}{\input{figures/exp_mnn.tex}}
  \caption{MNN topology used in experiments}
  \label{fig:exp_mnn}
\end{figure}

The Adaptive Moment Estimation (Adam) \cite{adam} has been used to compute adaptive learning rates for each parameter of the gradient descent optimization algorithms, carried out
with batch method. A learning rate of 0.001 has been set. The training has been carried on for 1000 epochs.

\begin{figure}[H]
\subfloat[Moons]{\label{fig:moons}\includegraphics[scale=.6]{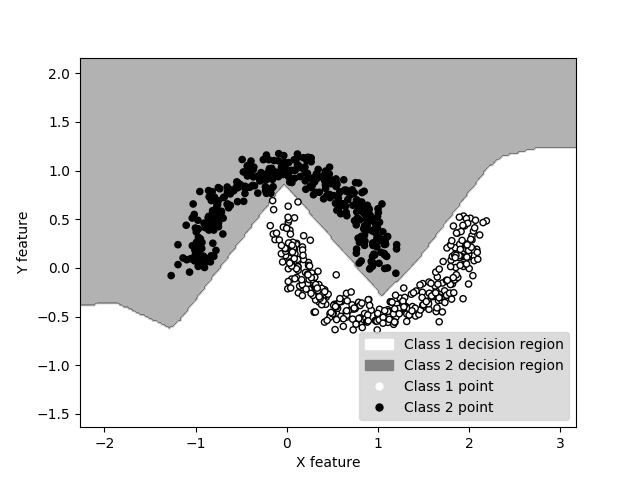}} \\
\subfloat[Circles]{\label{fig:circles}\includegraphics[scale=.6]{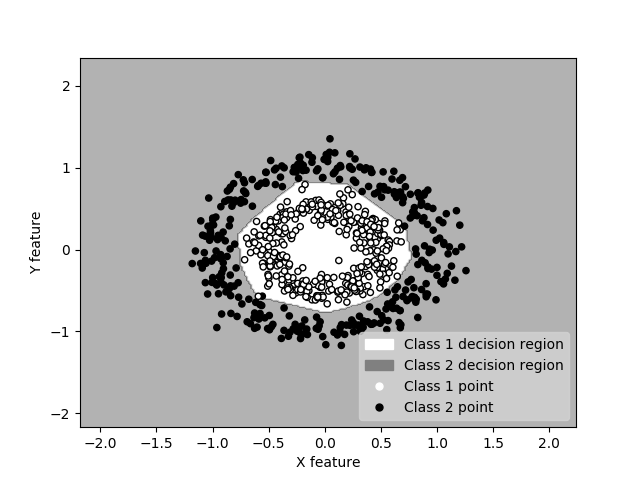}} \\
\subfloat[Spirals]{\label{fig:spirals}\includegraphics[scale=.6]{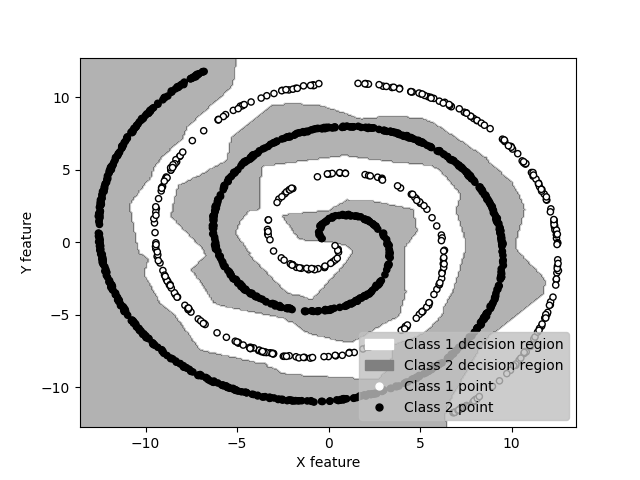}}
\caption{Two-classes datasets and related decision regions}
\label{fig:2_class_exp}
\end{figure}

\begin{figure}[!ht]
\subfloat[Single Blobs]{\label{fig:blobs}\includegraphics[scale=0.6]{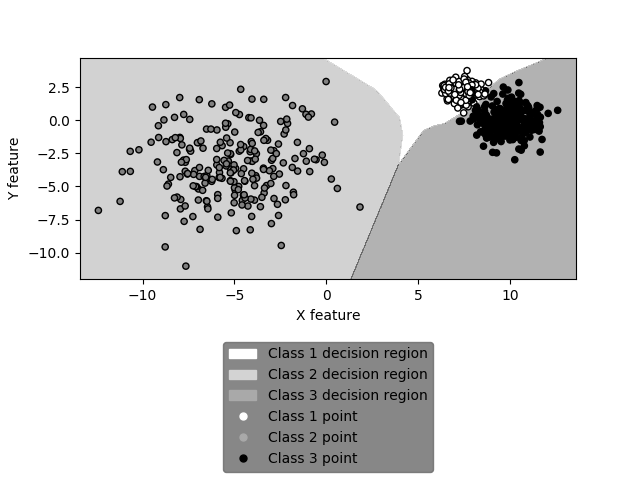}} \\
\subfloat[Double Blobs]{\label{fig:blobs2}\includegraphics[scale=0.6]{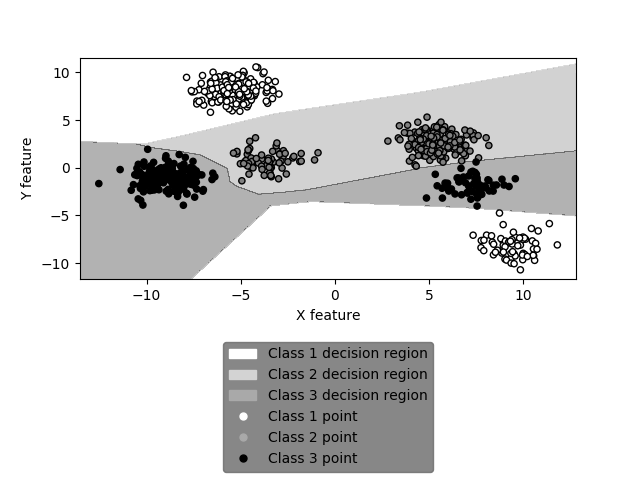}}
\caption{Three-classes datasets and related decision regions}
\label{fig:3_class_exp}
\end{figure}

The dataset has been partitioned into 70\% and 30\% for training and testing sets, respectively. Figures \ref{fig:2_class_exp}, \ref{fig:3_class_exp}, and \ref{fig:iris} show with different gray levels the resulting partitioning of the input domain made by the MNN. Here, the generalization capabilities of the network are apparent. As a result, the MNN achieved the 100\% accuracy for all datasets. In terms of complexity, the number of nodes of the MNN are $3+5+2=10$ and $3+5+3=11$ for 2 and 3 class datasets, respectively. The corresponding number of parameters (weights) is $10 \cdot 10=100$ and $11 \cdot 11=121$, respectively.
The interested reader is referred to \cite{mnnrepo} for a color animation of the MNN partitioning for each iteration.
Table \ref{fig:acc_hidd} shows the accuracy of the Spiral model generated by an MNN for increasing hidden neurons. It is interesting that, with 15 hidden neurons the problem is successfully modeled. Moreover, for a lower number of neurons, up to 7, the accuracy  decreases gradually, in contrast to MLP and other approaches proposed in \cite{Wilamowski2010}.

\begin{table}[!ht]
\centering
\begin{tabular}[t]{c|c}
Hidden Neurons & Accuracy   \\
\hline
5              & $0.75 \pm 0.079$ \\
7              & $0.95 \pm 0.029$ \\
10             & $0.94 \pm 0.039$ \\
13             & $0.95 \pm 0.026$ \\
15             & $0.99 \pm 0.011$
\end{tabular}
\caption{Accuracy of the Spiral model generated by an MNN for increasing hidden neurons}
\label{fig:acc_hidd}
\end{table}

Figure \ref{fig:iris}(a) and Figure \ref{fig:iris}(b) show the training loss and the training accuracy over time for the \textit{Iris} dataset. It is worth to note the convergence capabilities of the network.  
As a result, the MNN achieved $97.00\% \pm 1.62\%$ accuracy over 10 runs with a $3\sigma$ confidence interval.
In terms of complexity, the number of nodes of the MNN is $5+10+3=18$. The corresponding number of parameters (weights) is $18 \cdot 18=324$.

\begin{figure}[!ht]
\subfloat[Training loss over time]{\label{fig:loss}\includegraphics[scale=0.6]{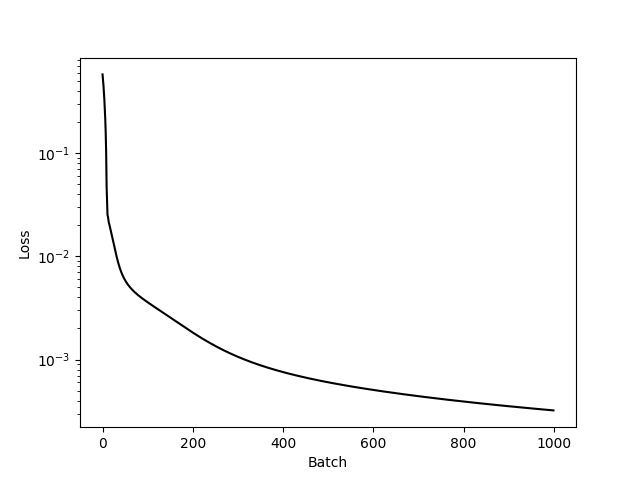}} \\
\subfloat[Training accuracy over time]{\label{fig:acc}\includegraphics[scale=0.6]{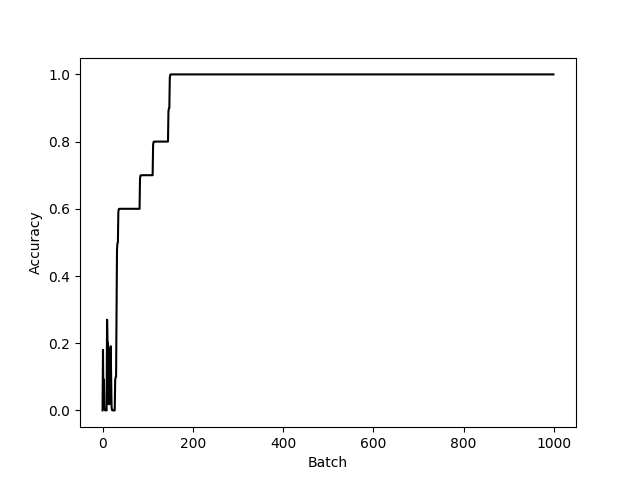}}
\caption{Training convergence of MNNs with Iris dataset}
\label{fig:iris}
\end{figure}

\clearpage

\subsection{Real-world problems}

\begin{figure}[!ht]
\subfloat[MNIST\label{fig:mnist}]{%
\begin{minipage}[c][1\width]{0.5\textwidth}%
\includegraphics[scale=0.5]{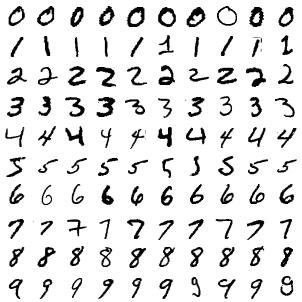}
\end{minipage}}
\subfloat[Fashion MNIST\label{fig:fashion}]{
\centering{}%
\begin{minipage}[c][1\width]{0.5\textwidth}%
\begin{center}
\includegraphics[scale=0.5]{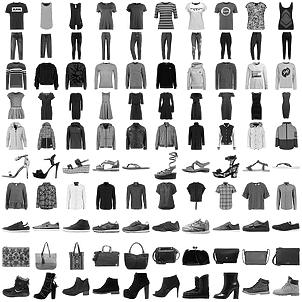}
\par\end{center}%
\end{minipage}}
\caption{Representative samples of MNIST and Fashion MNIST  datasets}
\label{fig:conv_datasets}
\end{figure}

To investigate the effectiveness of the MNN architecture, some experiments have been carried out on two real-world problems used for benchmarking machine learning algorithms: MNIST \cite{mnist} and Fashion-MNIST \cite{fashionmnist}. MNIST is a database of handwritten digits images, whereas Fashion-MNIST is a dataset of fashion article images. Both datasets are made by a training set of 60,000 examples, and a test set of 10,000 examples. Each example is a 28x28 image, with pixels in 0-255 grayscale values, associated with a class label of 10 possible classes. The task is to classify a given image into one of such 10 classes. Figure \ref{fig:conv_datasets} shows some representative samples of the datasets. Both datasets contain samples ambiguous even for humans: MNIST and Fashion-MNIST have an average human performance of 98.29\% \cite{NIPS1992} and 83.5\% \cite{fashionmnist}, respectively. Such datasets are widely used and deeply investigated: top-performing models, based on convolutional neural networks, achieve a classification accuracy higher than 99\%, and have a layered structure made by feature extraction and classification. Feature extraction can be performed by alternating convolution and subsampling layers, whereas classification can be performed via dense layers, such as a fully connected feed forward (i.e, MLP-based) neural network. 
The purpose of this section is to use an MNN as a classification layer, to carry out a comparative analysis between MNN and MLP. Rather than providing the top performance, this solution simplifies the design of the classification layer for the sake of simplicity. Indeed, for a fair comparison it is essential to avoid complex architectures with many hyper-parameters, whose particular choices should be subject to in-depth discussion. Similarly, there are many choices for convolutional architectures, but using a general-purpose architecture with a high degree of automation reduces such choices.

With this premise, a Convolutional Auto-Encoder (CAE) is used for feature extraction, followed by an MNN or MLP based network for classification. The CAE is commonly used for unsupervised data encoding and noise reduction \cite{convautoencoder}.
The following architecture is used in experiments. \textit{Encoding}: a convolutional layer with a 3x3 kernel size and 16 channels, stride 3 and padding 1; a rectified linear unit (ReLU); a max pooling layer with 2x2 kernel size, stride 2; a convolutional layer with a 3x3 kernel size and 8 channels, stride 2 and padding 1; a ReLU; a max pooling layer with 2x2 kernel size, stride 2. \textit{Decoding}: a transpose convolutional layer with 3x3 kernel size and 16 channels, stride 2; a ReLU; a transpose convolutional layer with 5x5 kernel size and 8 channels, stride 3, padding 1; a ReLU; a transpose convolutional layer with 2x2 kernel size and 1 channel, stride 2, padding 1; a hyperbolic tangent activation function.
Overall, the CAE provides 32 features to the classification layer.
Both MLP and MNN classification layers have been equipped with $\ninputs=32$ input and $\noutputs=10$ output neurons. The number $\nhiddens$ of hidden neurons has been set accordingly, to have the same number of overall connections for the two comparative networks. Since both datasets are spatial, in the MNN network two recursion steps, i.e., $\nticks = 3$ , are sufficient. The MNN network is statically pruned for better efficiency. Since the MNN model generalizes the other perceptron-based topologies, there are custom pruning that makes the MNN fully equivalent, for instance, to an RNN or to an MLP. However, for a significant comparison, such custom pruning is avoided, in favor of a randomly determined pruning.

\begin{figure}[!ht]
\includegraphics[scale=0.3]{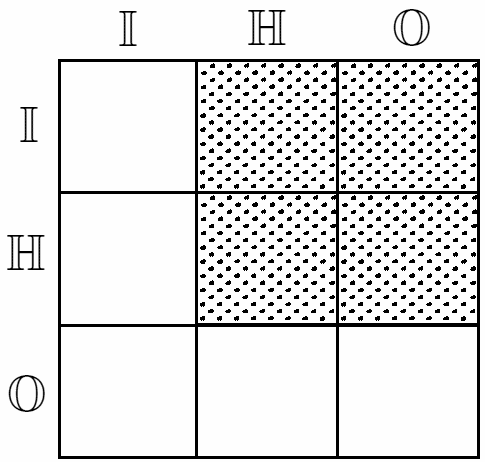}
\caption{Pruned adjacency matrix}
\label{fig:pruned_matrix}
\end{figure}

Figure \ref{fig:pruned_matrix} represents the adjacency matrix of the MNN based classification layer. Here,
$\gset{I}$, $\gset{H}$, and $\gset{O}$, represent the sets of indexes corresponding to the input, hidden and output neurons. Each block can then be characterized by a pair of related sets. In particular, white blocks have zero connections, whereas dotted blocks have a given percentage of randomly selected connections.

Specifically, the white blocks represent the following connection types: all-to-input ($\gset{I} \rightarrow \gset{I}$, $\gset{H} \rightarrow \gset{I}$, $\gset{O} \rightarrow \gset{I}$), output-to-all ($\gset{O} \rightarrow \gset{I}$, $\gset{O} \rightarrow \gset{H}$, $\gset{O} \rightarrow \gset{O}$). The dotted blocks represent the following connection types: input-to-hidden ($\gset{I} \rightarrow \gset{H}$), input-to-output ($\gset{I} \rightarrow \gset{O}$), hidden-to-hidden ($\gset{H} \rightarrow \gset{H}$), and hidden-to-output ($\gset{H} \rightarrow \gset{O}$). As a consequence, assuming $\nhiddens = 50$, the following connections and biases are available in the dotted area: $(1 - p) \cdot (\ninputs+\nhiddens) \cdot (\nhiddens + \noutputs) + (\nhiddens + \noutputs)$, where $p$ is the pruning percentage, and the last term $(\nhiddens + \noutputs)$ is the number of biases of hidden and output nodes. 
In order to have a similar number of connections, the MLP hidden neurons are made by two layers of $h_1$ and $h_2$ neurons. Hence, the total number of connections, considering also biases, is $(\ninputs \cdot h_1 + h_1) + (h_1 \cdot h_2 + h_2) + (h_2 \cdot \noutputs + \noutputs) $.
The 95\% confidence intervals achieved via the MNN and MLP based classifiers, calculated over 10 trials, are summarized in Table \ref{fig:conv_results}. It important to note that, for each trial, the percentage of randomly selected connections in MNN is completely renewed. As previously discussed, such classification rates are related to the features generated by the CAE layer. Consequently, the rates are not comparable with the top absolute performance of the literature. The significant result is that the MNN and the MLP based classifiers achieve very similar performance for increasing connections. For the sake of comparability, Table \ref{fig:conv_results} shows only the settings with a very similar number of connections for the two networks. The best performance, of about 0.80 classification rate, is achieved via 1044-1047 connections. To evaluate the complexity of the classification task, it has been experimentally verified that for increasing number of connections (up to more than 3 thousand connections), both classifiers are not able to overcome the 0.8 classification rate.

\begin{table}[!ht]
\centering
\begin{tabular}[t]{c|c|c|c|c|c}
Architecture & Hidden Neurons & Pruning & Connections & MNIST & Fashion MNIST \\
\hline
MNN &  50   & 85\% &  798  & $0.772 \pm 0.030$ & $0.762 \pm 0.005$ \\
MLP & 14+13 & -    &  797  & $0.761 \pm 0.065$ & $0.750 \pm 0.026$ \\
MNN &  50   & 80\% & 1044  & $0.796 \pm 0.013$ & $0.771 \pm 0.004$ \\
MLP & 17+17 & -    & 1047  & $0.789 \pm 0.035$ & $0.786 \pm 0.013$ 
\end{tabular}
\caption{Testing classification rates of the MLP and MNN based networks, on MNIST and fashion-MNIST}
\label{fig:conv_results}
\end{table}

\section{Conclusions and future work}
The purpose of this paper is to formally introduce recent advances leading to the MNNs, providing the key points to the reader.

Overall, the main advantages of the MNN model with the related FOP algorithm are: (i) the state partial derivatives can be computed along the forward propagation; (ii) the error gradient can be directly computed from state gradient; (iii) the state partial derivative update makes use only of short-lived variables, which can be overwritten at each state iteration; (iv) the state partial derivatives concern only one multidimensional parameter; (v) the overall gradient computation relies only on tensor multiplications, which can be easily distributed on parallel computing, thus potentially enabling large-scale sparse ANNs training \cite{matrixmul}. 

In contrast, the BP-based family of algorithms is limited to layer-wise architectures, and needs to store all intermediate layer outputs, by comprising a forward and backward propagation through the network. On the other side, the CG-based gradient computation is not constrained in terms of network architecture, but it needs to store a large graph topology and  the partial derivatives of each computation node, and it needs to compute all factoring paths for each parameter.

Due to its unconstrained structure, an interesting research perspective of MNNs is to adopt structural regularization techniques to dynamically drive the network topology.
For small datasets the network topology is highly dense to exploit the available neurons, and then the adjacency matrix is highly dense. For large datasets, in general there are two strategies that can make the adjacency matrix sparse: a) offline pruning, i.e., to remove some types of connections according to some heuristics; b) online pruning, i.e., to remove iteratively some connections that do not contribute to model, in the training phase. The two strategies can be combined. In general, the possibility to have a sparse matrix depends on the problem complexity. Since the MNN needs large matrix operations, such strategies should also be supported by a framework implementation that efficiently exploits the hardware resources, e.g. via memory caching and highly parallel computation. However, the commercially available machine learning framework provide optimized libraries for back-propagated models. As a consequence, to test the MNN architecture on very large datasets, an optimized framework should be implemented on specific hardware. Such development is a long-term task and it is out of the scope of this paper, which focuses on the formal derivation of the technique and on pilot experimentation showing its potential application.

As a future work, in order to compare BP, CG and FOP according to a performance perspective, the scalability of each algorithm should be evaluated in terms of computational complexity. Moreover, a statistical performance evaluation should be carried out on benchmark problems, considering large-scale applications. 

\section*{Acknowledgements}
This research was partially carried out in the framework of the following projects: (i) PRA 2018\_81 project entitled “Wearable sensor systems:
personalized analysis and data security in healthcare”
funded by the University of Pisa; (ii) CrossLab project (Departments of Excellence), funded by the Italian Ministry of Education and Research (MIUR); (iii) “KiFoot: Sensorized footwear for gait analysis” project, co-funded by the Tuscany Region (Italy) under the PAR FAS 2007-2013 fund and the FAR fund of the Ministry of Education, University and Research (MIUR).

\bibliography{bibliography} 
\bibliographystyle{ieeetr}

\end{document}

%% file: figures/ff.tex
\begin{tikzpicture}[
    neuron/.style={
        circle,
        draw,
        minimum height =0.7cm,
    }
]
    \pgfmathsetmacro\N{3}
    \pgfmathsetmacro\M{3}
    \pgfmathsetmacro\layers{2}
    
    \pgfmathsetmacro\offset{(\N-\M)/2}
    
    \pgfmathsetmacro\dist{1.5}
    \pgfmathtruncatemacro\lpo{\layers + 1}
    \foreach \layer in {0,...,\lpo}{
        \foreach \i in {0,...,\N}{
            \pgfmathtruncatemacro\l{\N - \i};
            \pgfmathsetmacro\npos{\dist * \layer};
            \ifnum \i=1
                \node (\layer\i) at (\npos,\i)  [] {\vdots};
            \else
                \node (\layer\i) at (\npos,\i)  [neuron] {};
            \fi
        }
    }
    
    \foreach \layer in {0,...,\layers}{
        \pgfmathtruncatemacro\nextlayer{\layer + 1} 
        \pgfmathsetmacro\labelpos{\layer*\dist + \dist/2}
        \node (label\layer) at (\labelpos, 3.5) [] {$W_\layer$};
        \foreach \i in {0,...,\N}{
            \foreach \j in {0,...,\M}{
                \ifnum \i=1
                \else
                    \ifnum \j=1
                    \else
                        \draw [-] (\layer\i) -- (\nextlayer\j) node[midway, above] {};
                    \fi
                \fi

            }
        }
    }

\end{tikzpicture}

%% file: figures/adj_ff.tex
\resizebox{.8\hsize}{!}{
$
    \begin{bmatrix}
        \begin{array}{c|c|c|c}
            0 & W_0 & 0 & 0\\
            \hline
            0 & 0 & W_1 & 0\\
            \hline
            0 & 0 & 0 & W_2\\
            \end{array}
    \end{bmatrix}
$
}

%% file: figures/am.tex
\begin{tikzpicture}[
    neuron/.style={
        circle,
        draw,
        minimum height =0.7cm,
    },
    ]
    \foreach [count=\i from 0] \coord in {(0,2),(0,1),(1.650640935899262,1.6666082764868995),(1.4238111791059187,0.8420470893021784),(3.460282475563993,0.45076567923470057),(3.157041715464337,1.5091140377585883),(1.496311214938192,2.50314239335771),(3.778845525166617,2.56042256996626294),(5,2),(5,1)}{
        \node[neuron] (n\i) at \coord {$n_{\i}$};
    }

    \foreach [count=\r from 0] \row in {{0,0,1,1,0,0,0,0,1,0},{0,0,0,1,0,1,1,0,0,0},{0,0,0,0,0,1,0,0,0,0},{0,0,0,0,0,0,0,0,0,1},{0,0,0,0,0,0,0,1,0,0},{0,0,0,0,0,0,0,0,0,1},{0,0,0,0,1,0,0,0,1,0},{0,0,1,0,0,0,0,0,1,1},{0,0,0,0,0,0,0,0,0,0},{0,0,0,0,0,0,0,0,0,0}}{
        \foreach [count=\c from 0] \cell in \row{
            \ifnum\cell=1
                \draw [->] (n\r) -- (n\c);
            \fi
        }
    }
\end{tikzpicture}

 

%% file: figures/adj_am.tex
\resizebox{1\hsize}{!}{
$
\begin{bmatrix}
    0 & 0 & w_{0,2} & w_{0,3} & 0 & 0 & 0 & 0 & w_{0,8} & 0 \\
    0 & 0 & 0 & w_{1,3} & 0 &  w_{1,5} &  w_{1,6} & 0 & 0 & 0 \\
    0 & 0 & 0 & 0 & 0 &  w_{2,5} & 0 & 0 & 0 & 0 \\
    0 & 0 & 0 & 0 & 0 & 0 & 0 & 0 & 0 &  w_{3,9} \\
    0 & 0 & 0 & 0 & 0 & 0 & 0 &  w_{4,7} & 0 & 0 \\
    0 & 0 & 0 & 0 & 0 & 0 & 0 & 0 & 0 &  w_{5,9} \\
    0 & 0 & 0 & 0 &  w_{6,5} & 0 & 0 & 0 & w_{6,8} & 0 \\
    0 & 0 & w_{7,2} & 0 & 0 & 0 & 0 & 0 & w_{7,8} & w_{7,9} \\
    0 & 0 & 0 & 0 & 0 & 0 & 0 & 0 & 0 & 0 \\
    0 & 0 & 0 & 0 & 0 & 0 & 0 & 0 & 0 & 0    
\end{bmatrix}
$
}

%% file: figures/am_algo.tex
\begin{algorithm}[H]
    \SetAlgoLined
    $\nabla_\gmatrix{A} \gvector{S} \gets 0$\\
    \For{i \textbf{in} $\{1,2,\cdots,\nticks-1\}$}{
        $\gvector{S}_{0:\ninputs} \gets x$\\
        $t \gets \gvector{S} \; \gmatrix{A}$\\
        $\nabla_\gmatrix{A} \gvector{S} \gets \nabla_\gvector{t}\varphi(\gvector{t}) \odot (\nabla_\gmatrix{A} \gvector{S} \; \gmatrix{A} + \gtensor{\tilde{S}} ) $ \\
        $\gvector{S} \gets \varphi(\gvector{t})$\\
    }
    $\gvector{y} \gets \gvector{s}[\nneurons-\nhiddens:\nneurons]$ \\
    $\nabla_\gmatrix{A} E(\gvector{y},\overline{\gvector{y}}) \gets \nabla_\gvector{y}E(\gvector{y},\overline{\gvector{y}})   \odot \nabla_\gmatrix{A} \gvector{S}$ \\
\end{algorithm}

%% file: figures/exp_mnn.tex
\begin{tikzpicture}[
    neuron/.style={
        circle,
        draw,
        minimum height =0.7cm,
    },
    ]
    \foreach [count=\i from 0] \coord in {(0,0),(0,4),(0,8),(5.738751462412286,6.728789164055736),(7.5545977258436,0.4716688167051659),(4.749246604838028,1.3792998454036742),(3.5370602699355587,3.858976050145034),(3.240040045246367,8.697448104695233),(10,0),(10,4),(10,8)}{
        \node[neuron] (n\i) at \coord {$n_{\i}$};
    }

    \foreach [count=\r from 0] \row in {{0,0,0,1,1,1,1,1,1,1,1},{0,0,0,1,1,1,1,1,1,1,1},{0,0,0,1,1,1,1,1,1,1,1},{0,0,0,0,1,1,1,1,1,1,1},{0,0,0,1,0,1,1,1,1,1,1},{0,0,0,1,1,0,1,1,1,1,1},{0,0,0,1,1,1,0,1,1,1,1},{0,0,0,1,1,1,1,0,1,1,1},{0,0,0,1,1,1,1,1,0,0,0},{0,0,0,1,1,1,1,1,0,0,0},{0,0,0,1,1,1,1,1,0,0,0}}{
        \foreach [count=\c from 0] \cell in \row{
            \ifnum\cell=1
                \draw [->] (n\r) -- (n\c);
            \fi
        }
    }
\end{tikzpicture}